\documentclass[lettersize,journal]{IEEEtran}
\usepackage{amsmath,amsfonts}
\usepackage{algorithmic}
\usepackage{algorithm}
\usepackage{array}
\usepackage[caption=false,font=normalsize,labelfont=sf,textfont=sf]{subfig}
\usepackage{textcomp}
\usepackage{stfloats}
\usepackage{url}
\usepackage{verbatim}
\usepackage{graphicx}
\usepackage[style=ieee]{biblatex}
\usepackage{xcolor}
\usepackage{amssymb}
\usepackage{pgfplots}
\usepackage{tikz}

\definecolor{bblue}{HTML}{4F81BD}
\definecolor{rred}{HTML}{C0504D}
\definecolor{ggreen}{HTML}{9BBB59}
\definecolor{ppurple}{HTML}{9F4C7C}
\newcommand{\ie}{\emph{i.e.}~}

\bibliography{main.bib}{}

\begin{document}

\title{TriMix: Virtual embeddings and self-consistency for self-supervised learning}

\author{Tariq Bdair, Hossam Abdelhamid, Nassir Navab, and Shadi Albarqouni
        % <-this % stops a space
\thanks{T. Bdair, H. Abdelhamid, N. Navab, and S. Albarqouni are with the Chair for Computer Aided Medical Procedure, Technical University of Munich, 85748 Munich, Germany (\textit{Corresponding author} e-mail: t.bdair@tum.de).}
\thanks{N. Navab is also with the Whiting School of Engineering, Johns Hopkins University, Baltimore, MD 21218 USA.}
\thanks{S. Albarqouni is also with  Clinic for Diagnostic and Interventional Radiology, University Hospital Bonn, Germany.}
\thanks{S. Albarqouni is also with  Helmholtz AI, Helmholtz Zentrum München, Germany.}
}

\markboth{}%
{Author \MakeLowercase{\textit{et al.}}: TriMix}

\maketitle

\begin{abstract}
Self-supervised Learning (SSL) has recently gained much attention due to the high cost and data limitation in the training of supervised learning models. The current paradigm in the SSL is to utilize data augmentation at the input space to create different views of the same images and train a model to maximize the representations between similar images and minimize them for different ones. While this approach achieves state-of-the-art (SOTA) results in various downstream tasks, it still lakes the opportunity to investigate the latent space augmentation. This paper proposes TriMix, a novel concept for SSL that generates virtual embeddings through linear interpolation of the data, thus providing the model with novel representations. Our strategy focuses on training the model to extract the original embeddings from virtual ones, hence, better representation learning. Additionally, we propose a self-consistency term that improves the consistency between the virtual and actual embeddings. We validate TriMix on eight benchmark datasets consisting of natural and medical images with an improvement of $2.71\%$ and $0.41\%$ better than the second-best models for both data types. Further, our approach outperformed the current methods in semi-supervised learning, particularly in low data regimes. Besides, our pre-trained models showed better transfer to other datasets. 
\end{abstract}


\section{Introduction}
\label{intro}
The scarcity of labeled data is one of the most challenging problems in achieving reliable performance in deep learning methods. Yet, current self-supervised learning (SSL) approaches have shown remarkable advances in downstream tasks such as computer vision \cite{bachman2019learning,bardes2021vicreg,cao2020parametric,caron2020unsupervised,chen2020simple,chen2020big,chen2021exploring,grill2020bootstrap,he2020momentum,srinivas2020curl,tian2021divide,tian2020contrastive,zbontar2021barlow}, natural language processing \cite{devlin2018bert,yang2019xlnet}, and speech recognition \cite{oord2018representation,xiong2016achieving}. SSL methods do not rely on a massive amount of annotated data. Instead, they train a model to produce good representations of the unsupervised data, a.k.a pretext task that help in a supervised task such as image classification and segmentation, a.k.a downstream task.   
An evolving direction in SSL methods known as contrastive learning utilizes siamese networks~\cite{bromley1993signature} to maximize agreement between different views of the same image, known as positive samples while decreasing it with other images, \ie the negative examples as proposed in SimCLR~\cite{chen2020simple} and MoCO~\cite{he2020momentum}.   
One drawback of the previous works requires expensive computations to find the negative images from a memory bank~\cite{he2020momentum} or a large batch size~\cite{chen2020simple}. SwAV~\cite{caron2020unsupervised} overcomes this limitation by clustering the samples based on the similarities of their features while forcing a consistency between cluster assignments produced for the positive examples.
Yet, BYOL~\cite{grill2020bootstrap} and SimSiam~\cite{chen2021exploring} relax the necessity for negative samples by employing asymmetric network tricks to avoid model failure while achieving state-of-the-art results. 
Recent methods were proposed based on information maximization to avoid features collapse via (i) whiting approaches; W-MSE~\cite{cao2020parametric}, (ii) redundancy reduction; Barlow Twins~\cite{zbontar2021barlow}, (iii) features decorrelation and normalization; Shuffled-DBN~\cite{hua2021feature}, and (iv) variance-preservation term; VICReg~\cite{bardes2021vicreg}.
While the previous methods work properly on highly curated datasets for pretraining such as ImagNet~\cite{deng2009imagenet}, DnC~\cite{tian2021divide} alternates between the contrastive learning and clustering-based methods to improve the performance on less curated datasets.

So far, the above methods utilize augmentation at the input space to create different views of the same image to learn better representations. However, a couple of non-self-supervised works have shown a boost in performance in image classification~\cite{berthelot2019mixmatch,verma2019manifold} or medical image segmentation~\cite{bdair2020roam,chaitanya2019semi,eaton2018improving,jung2019prostate,panfilov2019improving} via the augmentation at the input space~\cite{berthelot2019mixmatch,eaton2018improving,chaitanya2019semi,panfilov2019improving}, the hidden representations~\cite{verma2019manifold}, or randomly at the input and hidden layers~\cite{bdair2020roam,jung2019prostate}. 
Although the latter process provides virtual data points that are beneficial to the model during the training, none of the former self-supervised learning methods has investigated that. Therefore, we propose our method that provides the model with virtual embeddings created at the hidden layers to learn better representations.   

In this paper, we propose \textbf{TriMix}, a novel concept for self-supervised learning that leverages the virtual data augmentation at the input and hidden embeddings, at the same time forcing the model to predict the percentage of such compositions in the virtual data from its original ones. Our method performs a linear interpolation on the input images and their corresponding hidden embeddings. Likewise, the mixed samples are fed to the model to generate virtual features. During the training, the model learns to decompose the mixed-up augmented features to their original components through our \textbf{\textit{virtual embeddings loss}}.
Note that the virtual embeddings are generated from a series of non-linear operations of the mixed-up data at the network. At the same time, the mixed-up data is produced from the linear process of the input images. Thus, matching the mixed-up data with its virtual embeddings might not be straightforward. To resolve this issue, we propose our \textbf{\textit{self-consistency loss}} which ensures the linearity and forces both embeddings to be consistent. To this end, our \textbf{contributions} are:
\begin{itemize}
\item We propose \textbf{TriMix}, a novel method for self-supervised learning that leverages the augmentation at hidden embeddings in training and guides the model to decompose the mixed data to its original components through our \textbf{\textit{virtual embeddings loss}}. Furthermore, the newly generated representations are fine-tuned via redundancy reduction techniques to learn better discriminative features.        
\item We propose \textbf{\textit{self-consistency loss}} to force the linearity and consistency between virtual embeddings and mixed-up embeddings for better training.
\item We compare \textbf{TriMix} with recent self-supervised learning methods on eight benchmark datasets of natural and medical datasets while showing superior performance.
\end{itemize}

\section{Related Work}
\label{related}
\subsection{Contrastive learning methods}
These methods, \emph{e.g.} \cite{chen2020simple,he2020momentum,henaff2020data,hjelm2018learning,oord2018representation,wu2018unsupervised}, learn valuable representations by minimizing the distance between similar views, a.k.a positive samples, generated from the same input while maximizing it with not alike ones, a.k.a negative examples. The views can be generated, for instance, using data augmentation methods such as color decomposition, random cropping (with flip and resize), Gaussian blur, and color distortion. However, these methods depend on InfoNCE loss~\cite{oord2018representation}, siamese networks~\cite{bromley1993signature}, and many negative examples to perform well. Although these approach have reduced the gap with supervised learning, finding the negative examples is an expensive procedure.
\subsection{Clustering methods} 
Clustering-based methods overcome the necessity of distinguishing between individual samples by differentiating between groups of images clustered based on their likenesses~\cite{asano2019self,caron2018deep,caron2020unsupervised,xie2016unsupervised}. For instance, DeepCluster~\cite{caron2018deep} utilized K-mean assignments as priors to cluster the learned representations. On the other hand, SwAV~\cite{caron2020unsupervised} applied an online clustering approach while forcing agreement between the representations from several views of the same image. Nonetheless, these methods require a lot of negative samples to produce reliable predictions.
\subsection{Asymmetric architectures methods} 
In a different work direction,  \cite{chen2021exploring,grill2020bootstrap} trained self-supervised models without relying on the negative examples. The idea is to have a siamese architecture and online and target networks. The online network, with learnable parameters, is trained to predict the presentations for the target network. BYOL~\cite{grill2020bootstrap}, for example, sets the parameters of the target network as moving average parameters of the online network. However, the parameters in SimSiam~\cite{chen2021exploring} are shared between both networks, while backpropagation and stop-gradient trick are applied to online and target networks, respectively. Despite that these tricks avoid the collapse solutions, they lack the explainablty~\cite{bardes2021vicreg}.
\subsection{Information maximization methods} 
An elegant method, Barlow Twins~\cite{zbontar2021barlow}, utilizes the redundancy-reduction principle to make the cross-correlation matrix, produced from two siamese features, close to the identity matrix. W-MSE~\cite{cao2020parametric} achieves this by whitening feature representations within each batch via Cholesky decomposition. To prevent a dimensional collapse, ~\cite{hua2021feature} proposed shuffled de-correlated batch normalization (DBN)~\cite{huang2018decorrelated}. VicReg~\cite{bardes2021vicreg} proposed another method free from the normalization step via employing Variance-Invariance-Covariance regularization terms.  

\section{TriMix}
\label{method}
\begin{figure*}[!t]
\centering
\includegraphics[scale=0.15]{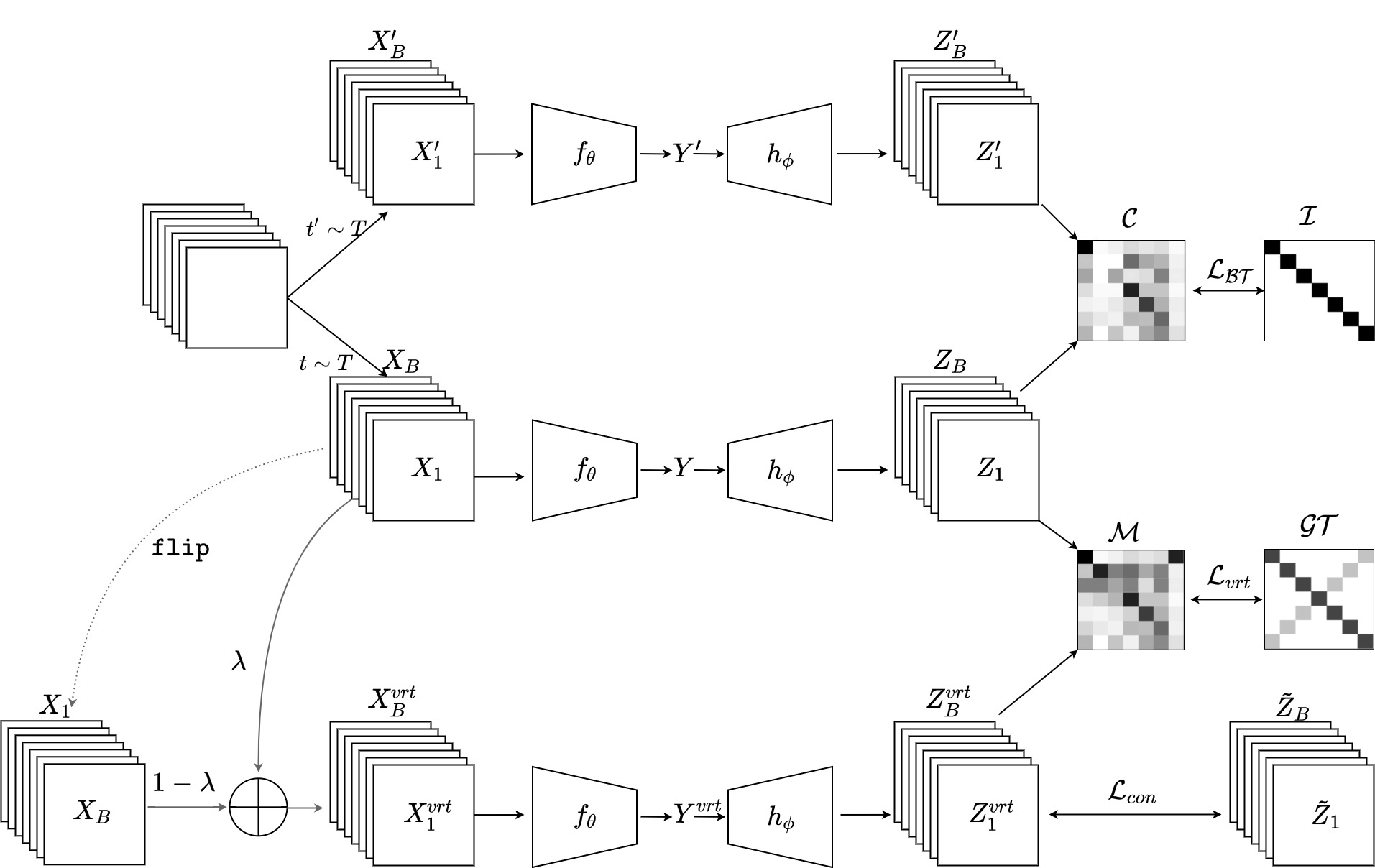}
\caption{\textbf{TriMix} mainly consists of virtual embeddings and self-consistency. Virtual embeddings: First, virtual data $X^{vrt}$ is created by linear interpolation of the input images. $X^{vrt}$ is then fed to the network to create the virtual embeddings $Z^{vrt}$. The model is trained to decompose the virtual data to the original ones using $\mathcal{L}_{vrt}$ loss, see Eq.(\ref{eq_L_vrt}). Self-consistency: to force the consistency between the virtual embeddings $Z^{vrt}$ and augmented embeddings $\Tilde{Z}$, we propose $\mathcal{L}_{con}$, see Eq.(\ref{eq_L_con}), where $\Tilde{Z}$ is created using Eq.(\ref{eq_mixupZ}). PyTorch alike pseudo code is presented in Algorithm \ref{alg:algo1}.}
\label{figTriMix}
\end{figure*}
Our method has the same architecture employed in the recent self-supervised learning methods~\cite{bardes2021vicreg,cao2020parametric,chen2021exploring,hua2021feature,grill2020bootstrap,zbontar2021barlow} where a siamese network~\cite{bromley1993signature} is trained on joint embedding on distorted images. In addition, our strategy proposes a mutual training of redundancy reductions and latent space augmentation approaches. Thus, we build upon information maximization approaches to include our contributions. 
Specifically, we borrow the redundancy-reduction principle from Barlow Twins~\cite{zbontar2021barlow} before combining it with augmented embeddings and self-consistency methods.
\subsection{Background}
Given a batch of input images $I$ sampled from a dataset $\mathcal{D}$, two different views $(X, X^\prime)$ are generated by applying two transformations $t$ and $t^\prime$ on $I$, where $X=t(I)$, $X^\prime=t^\prime(I)$, and $t$ and $t^\prime$ are sampled from a distribution of data augmentations ${T}$. Then, $X$ and $X^\prime$ are encoded to a deep neural network with trainable parameters $f_{\theta}$ to produce the hidden representations $Y=f_{\theta}(X)$ and $Y^\prime=f_{\theta}(X^\prime)$, respectively.  Next, these representations are fed to projector $h_{\phi}$ to create the embeddings $Z$ and $Z^\prime$, where $Z=h_{\phi}(Y)$ and $Z^\prime=h_{\phi}(Y^\prime)$. The embeddings are then normalized along the batch dimension to produce unit vectors with $0$ mean. 

The training loss appeared in Barlow Twins~\cite{zbontar2021barlow}, consists of two terms; invariance $\mathcal{L}_{inv}$ and redundancy reduction $\mathcal{L}_{rr}$ terms as follows.   
\begin{equation} 
\label{eq_L_bt}
\mathcal{L_{BT}} \triangleq \mathcal{L}_{inv} + \alpha \mathcal{L}_{rr},
\end{equation}
where $\alpha$ is a hyperparameter, and $\mathcal{L}_{inv}$ and $\mathcal{L}_{rr}$ are given by \ref{eq_L_inv} and \ref{eq_L_rr} , respectively.    
\begin{equation} 
\label{eq_L_inv}
\mathcal{L}_{inv} = \sum_{i} (1 - \mathcal{C}_{ii})^{2}, 
\end{equation}
\begin{equation} 
\label{eq_L_rr}
\mathcal{L}_{rr} = \sum_{i} \sum _{i \neq i} {\mathcal{C}_{ij}}^{2},
\end{equation}
where $\mathcal{C}$ is the cross-coloration matrix between the two outputs of the network and given by  
\begin{equation} 
\label{eq_CorMatC}
\mathcal{C}_{ij} \triangleq \frac{\sum_{b}z_{b,i}\hspace{5pt}z^{\prime}_{b,j}}{\sqrt{\sum_{b}(z_{b,i})^{2}}\sqrt{\sum_{b}(z^{\prime}_{b,j})^{2}}},
\end{equation}
where $i,j$ index to the vector dimension of the networks' outputs, and $b$ indexes to the batch samples. Note that $\mathcal{C} \in \mathbb{R}^{d \times d}$ is a square matrix with a size equal to the output dimension, with entries between $(1)$ for perfect correlation and $(-1)$ for perfect anti-correlation.    
Barlow Twins' objective function tries to find the best representations that preserve as much information about the samples. At the same time, it is agnostic or less informative about the distortions applied to these samples.    
\subsection{Augmented Embeddings}
To augment the model with new data points, our method, shown in Fig.(\ref{figTriMix}), takes one view of the input, \ie $X$, and flips it to create a reversed version $X^{r}$, where $X^{r}=\verb!flip!(X)$. Then, virtual data is generated by applying linear interpolation, \ie Mixup~\cite{zhang2017mixup}, between the original and the reversed versions as follows.
\begin{equation} 
\label{eq_mixupX}
{X}^{vrt} =  \lambda*X+(1-\lambda)*X^{r},
\end{equation}
where the mixup factor $\lambda$ is randomly sampled from the $Uniform$ distribution; \ie $\lambda \in [0, 1]$. Note that the Mixup is performed on one arm of the siamese network with its reversed version to guarantee that no sample is mixed-up with itself. Once we generate the mixed-up data, we pass it to the model to produce virtual embeddings;  ${Z}^{vrt}=h_{\phi}(f_{\theta}({X}^{vrt})$. Then, ${Z}^{vrt}$ is normalized along the features dimension and batch size to produce unit vectors with $0$ mean.
\subsection{Embeddings Decomposition}
We train the model to decompose the virtual embeddings to their original components to learn useful representations from the new virtual data points and their embeddings. In other words, we train the model to predict the values of the mixup factor $\lambda$. To achieve this, first, we create a cross-correlation matrix computed between the original embeddings of the input $X$ and virtual embeddings, \ie $(Z, {Z}^{vrt})$. 
\begin{equation} 
\label{eq_CorMatM}
\mathcal{M}_{mn} \triangleq \frac{\sum_{a}z_{a,m}\hspace{5pt}{z}^{vrt}_{a,n}}{\sqrt{\sum_{a}(z_{a,m})^{2}}\sqrt{\sum_{a}({z}^{vrt}_{a,n})^{2}}},
\end{equation}
where $m,n$ index to the batch samples, and $a$ indexes the embeddings' vector dimension. Note that $\mathcal{M} \in \mathbb{R}^{B \times B}$ is a square matrix with a size equal to the batch size $B$. To this end, each row represents the similarities between an image $m$ in the original batch and all the images in the virtual batch. Then, $\mathcal{M}$ is normalized using softmax operation to generate distributions with probabilities between $[0, 1]$ along its rows.
\begin{equation}  
\label{eq_soft}
\mathcal{M}_{m,:} = \texttt{\textbf{softmax}}(\mathcal{M}_{m,:}/\tau),
\end{equation}
where $m$ indexes the batch sample, and $\tau$ is the temperature hyperparameter.\\

\noindent\textbf{\textit{Virtual embeddings loss.}} 
To enforce the model to regress the percentage of mixed-up data composition, we propose our virtual embeddings loss as the absolute mean difference between the matrix $\mathcal{M}$ and our ground truth matrix $\mathcal{GT}$, and given by 
\begin{equation}  
\label{eq_L_vrt}
\mathcal{L}_{vrt} = \| \mathcal{M} - \mathcal{GT} \|
\end{equation}
where $\mathcal{GT} =\lambda I+(1-\lambda)(I*R), \mathcal{GT} \in \mathbb{R}^{B \times B}$, where $I \in \mathbb{R}^{B \times B}$ is a square identity matrix with a size equal to the batch size $B$, and $R$ is a transformation matrix that rotates $I$ counterclockwise by $90$ degree, see $\mathcal{GT}$ in Fig.(\ref{figTriMix}). 

\subsection{Self Consistency}
Thus far, we train the model to decompose the virtual embeddings to their original components. However, the mixed-up data ${X}^{vrt}$ are generated from linear interpolation of the input images $(X, {X}^{r})$. In contrast, the virtual embeddings ${Z}^{vrt}$ are generated from a series of non-linear operations of ${X}^{vrt}$ at the network. Thus, training the model to predict this linear operation of the mixed-up data from non-linear virtual embeddings might be challenging. To resolve this issue, we force the linearity and consistency between the virtual embeddings ${Z}^{vrt}$ and \textit{mixed-up embeddings}; $\Tilde{Z}$. Such that we define $\Tilde{Z}$ as the result of linear interpolation of the original inputs embeddings and their reversed version $(Z, Z^{r})$, and given by 
\begin{equation}
\label{eq_mixupZ}
\Tilde{Z} =  \lambda*Z+(1-\lambda)*Z^{r},
\end{equation}
where $Z^{r}=\verb!flip!(Z)$, and $\lambda$ is the same one used in Eq.(\ref{eq_mixupX}). \\

\noindent\textbf{\textit{Self-consistency loss.}}
Consequently, we propose our loss as the mean absolute difference between the two embeddings and the mixed-up embeddings to force the linearity and consistency between the virtual embeddings.  
\begin{equation}
\label{eq_L_con}
\mathcal{L}_{con} = \| \Tilde{Z} - {Z}^{vrt}\|.
\end{equation}
\subsection{Overall objective}
The overall objective function is the sum of Barlow Twins, virtual embeddings, and self-consistency losses, and given by
\begin{equation}
\label{eq_L_over}
\mathcal{L} =  \mathcal{L_{BT}} + \beta \mathcal{L}_{vrt} + \gamma \mathcal{L}_{con}
\end{equation}
where $\beta$ and $\gamma$ are hyperparameters. 
\begin{algorithm}[!t]
\fontsize{8pt}{8pt}
\caption{PyTorch-style pseudocode for TriMix}
\label{alg:algo1}

\begin{algorithmic}[1]

\STATE {\color{teal}{\# \texttt{net: f$_{\theta}(h_{\phi}())$}}}

\STATE {\color{teal}{\# \texttt{$\alpha$,$\beta$,$\gamma$: hyperparameters}}}

\STATE {\color{teal}{\# \texttt{uniform: uniform distribution}}}

\STATE {\color{teal}{\# \texttt{B: batch size}}}

\STATE {\color{teal}{\# \texttt{D: embeddings dimensionality}}}

\STATE {\color{teal}{\# \texttt{mm: matrix-matrix multiplication}}}

\STATE {\color{teal}{\# \texttt{eye: identity matrix}}}

\STATE {{\texttt{\textbf{for tuple in dataloader:}}}}

\STATE {\quad \texttt{\textbf{x, x$^{\prime}$ = tuple}}\color{teal}{ \# \texttt{sample two augmented views}}}

\STATE {\quad \texttt{\textbf{z, z$^{\prime}$ = net(x), net(x$^{\prime}$)}}\color{teal}{ \# \texttt{produce embeddings}}}

\STATE {\quad\color{teal}{\# \texttt{Barlow Twins}}}

\STATE {\quad \texttt{\textbf{z = (z - z.mean(0))/z.std(0)}}\color{teal}{ \#\texttt{BxD}}}

\STATE {\quad \texttt{\textbf{z$^{\prime}$ = (z$^{\prime}$ - z$^{\prime}$.mean(0))/z$^{\prime}$.std(0)}}\color{teal}{ \#\texttt{BxD}}}

\STATE {\quad\color{teal}{\# \texttt{DxD cross-correlation matrix}}}

\STATE {\quad \texttt{\textbf{c = mm(z.T, z$^{\prime}$)/B}}}

\STATE {\quad\color{teal}{\# \texttt{loss calculation for Barlow Twins}}}

\STATE {\quad \texttt{\textbf{c\char`_diff = (c-eye(D)).pow(2)}}}

\STATE {\quad \texttt{\textbf{off\char`_diagonal(c\char`_diff).mul\char`_($\alpha$)}}}

\STATE {\quad \texttt{\textbf{$\mathcal{L_{BT}}$ = c\char`_diff.sum()}}}

\STATE {\quad\color{teal}{\# \texttt{TriMix: 1. Augmented Embeddings}}}

\STATE {\quad \texttt{\textbf{$\lambda$= uniform()}}\color{teal}{ \# \texttt{sample the mixing factor}}}

\STATE {\quad \texttt{\textbf{x$^{r}$= flip(x)}}\color{teal}{\# \texttt{reversed version}}}

\STATE {\quad \texttt{\textbf{z$^{r}$= flip(z)}}\color{teal}{\# \texttt{reversed embeddings}}}

\STATE {\quad \texttt{\textbf{x$^{vrt}$ = $\lambda$*x + (1-$\lambda$)*x$^{r}$}}\color{teal}{ \# \texttt{mixed-up/virtual data}}}

\STATE {\quad \texttt{\textbf{z$^{vrt}$ = net(x$^{vrt}$)}}\color{teal}{ \# \texttt{virtual embeddings}}}

\STATE {\quad \texttt{\textbf{$\Tilde{\texttt{\textbf{z}}}$ = $\lambda$*z + (1-$\lambda$)*z$^{r}$}}\color{teal}{ \# \texttt{mixed-up  embeddings}}}

\STATE {\quad\color{teal}{\# \texttt{TriMix: 2. Features Decomposition}}}

\STATE {\quad\color{teal}{\# \texttt{Normalization along D and B}}}
\STATE {\quad \texttt{\textbf{z$^{vrt}$ = (z$^{vrt}$ - z$^{vrt}$.mean(0))/z$^{vrt}$.std(0)}}}

\STATE {\quad \texttt{\textbf{z$^{vrt}$ = ((z$^{vrt}$.T - z$^{vrt}$.mean(1))/z$^{vrt}$.std(1)).T}}}

\STATE {\quad \texttt{\textbf{m = mm(z, z$^{vrt}$.T)/D}}\color{teal}{ \# \texttt{BxB matrix}}}

\STATE {\quad \texttt{\textbf{m = softmax(m(0)/$\tau$)}}\color{teal}{ \# \texttt{softmax normalization}}}

\STATE {\quad\color{teal}{\# \texttt{Create our ground truth}}}

\STATE {\quad \texttt{\textbf{gt = $\lambda$*eye(B)+($1-\lambda$)*rotation90(eye(B))}}}

\STATE {\quad\color{teal}{\# \texttt{Virtual embeddings loss}}}
\STATE {\quad \texttt{\textbf{$\mathcal{L}_{vrt}$ = L1Loss(gt-m)}}}

\STATE {\quad\color{teal}{\# \texttt{TriMix: 3. Self Consistency}}}

\STATE {\quad \texttt{\textbf{$\mathcal{L}_{con}$ = L1Loss($\Tilde{\texttt{\textbf{z}}}$-z$^{vrt}$)}}}

\STATE {\quad\color{teal}{\# \texttt{Over all loss} \# }}

\STATE {\quad \texttt{\textbf{loss = $\mathcal{L_{BT}}$ + $\beta \mathcal{L}_{vrt}$ + $\gamma \mathcal{L}_{con}$}}}

\STATE {\quad \texttt{\textbf{loss.backward()}}}

\STATE {\quad \texttt{\textbf{optimizer.step()}}}

\end{algorithmic}

\end{algorithm}

\section{Experiments}
\label{exp}
\subsection{Experiments Setup}
\label{exp_setup}
\noindent\textbf{\textit{Datasets.}} We conduct our experiments on 8 public benchmarks. (i) \textbf{CIFAR-10}~\cite{krizhevsky2009learning} and (ii) \textbf{CIFAR-100}~\cite{krizhevsky2009learning}. Both datasets consists of $32\times32$ images with $10$ and $100$ classes, respectively. (iii) \textbf{STL10}~\cite{coates2011analysis}, consists of $96\times96$ images with $10$ classes. (iv) \textbf{Tiny ImageNet}~\cite{le2015tiny}, consists of $64\times64$ images with $200$ classes. Four medical datasets from \textbf{MedMNIST}\cite{medmnistv1,medmnistv2}. \textbf{MedMNIST} provides an MNIST-like set of standardized biomedical images consisting of 18 datasets with different scales and tasks. In this work, we randomly opt for four 2D multi-class datasets as follow. (i) \textbf{PathMNIST}: 107,180 colon pathology images of 9 classes. (ii) \textbf{DermaMNIST}: 10,015 dermatoscopic images of 7 classes. (iii) \textbf{OCTMNIST}:  109,309 retinal OCT images of 4 classes. (iv) \textbf{BloodMNIST}: 17,092 blood cell microscopic images of 8 classes. All the medical images are provided with the size of $28\times28$. \\

\noindent\textbf{\textit{Image augmentations.}} To produce the two views of the images, we follow the standard data augmentations used in the community. Specifically, random cropping, color jittering, horizontal flipping, and grayscaling were applied.\\

\noindent\textbf{\textit{Implementation Details.}} Adam optimizer~\cite{kingma2014adam} is utilized for training the models for 200 epochs, with a learning rate of $1\times10^{-3}$, weigh decay of $1\times10^{-6}$, and batch size of $256$. We adopt ResNet-18~\cite{he2016deep} as back-bone encoder with 512 output units. A 3-layer MLP with hidden layers of the size of $1024$ is used as a projector. All hidden layers are followed by the batch normalization layer and ReLU activation. We set $\alpha=5\times10^{-3}$ as in Barlow Twins~\cite{zbontar2021barlow}, and perform grid search for $\beta$ and $\gamma$ where it found best at $1000$ and $200$, respectively. $\tau$ set to $2$ as widely adopted in the literature.  
We opt for the PyTorch framework as an implementation environment hosted on a standalone NVIDIA Tesla V100 (Volta) with a 32 GB machine. The average training time takes around $7-8$ hours for each approach.  
\subsection{Results}
\label{exp_res}
\begin{figure*}[!t]
\centering
\includegraphics[width=\textwidth]{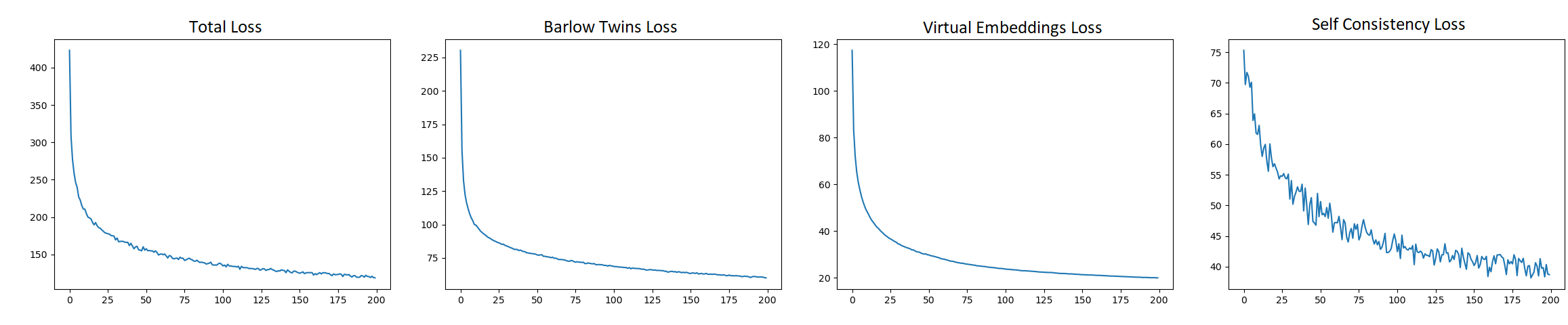}
\caption{TriMix training curves show that all losses converged and contributed significantly to the training on the TinyImage dataset. While tiny oscillations are noticed in the self-consistency loss, the overall trend is decreased.}
\label{figLoss}
\end{figure*}

\noindent\textbf{\textit{KNN and Linear Evaluations (Natural images).}}
We evaluate the pre-trained representations using a supervised linear classifier on the frozen representations following the standard procedures. Specifically, after an unsupervised pre-training on the training sets for 200 epochs, the features were frozen, then a supervised linear classifier, consisting of a fully-connected layer followed by a softmax layer, was trained on the extracted features for 100 epochs. We use a learning rate of $1\times10^{-3}$, weight decay of $1\times10^{-6}$, a momentum of $0.9$, and a batch size of $256$. The results are reported in Table~\ref{tablelinear}.
\begin{table*}[!t] 
\caption{Top-1 accuracy($\%$) of TriMix and the baselines for KNN (with 200-epoch pretraining) and linear evaluations ( with 100-epoch supervised training) on the benchmark datasets. All models use a ResNet-18 encoder and the same projector and augmentations. The best results are in bold. Our method outperforms other methods at different datasets in both KNN and linear evaluations.}
\label{tablelinear}
\centering
\resizebox{1\textwidth}{!}{
\begin{tabular}{l|l|c c c c| c c c c}
\hline
\multicolumn{2}{c}{}& \multicolumn{4}{c|}{\textbf{KNN}}&\multicolumn{4}{c}{\textbf{Linear}}\\
\hline
\multicolumn{2}{c|}{Method}&CIFAR10&CIFAR100&STL10&Tiny ImageNet& CIFAR10&CIFAR100&STL10&TinyImageNet\\
\hline 
\hline
\multicolumn{2}{c|}{SimCLR}&82.17&48.06&79.47&30.33&86.14&59.27&86.35&42.74\\
\multicolumn{2}{c|}{BYOL}&81.13&45.73&81.19&30.75&85.43&57.31&86.33&41.94\\
\multicolumn{2}{c|}{Barlow Twins}&84.37&52.47&80.98&\textbf{36.70}&86.93&60.66&86.29&\textbf{45.50}\\
\multicolumn{2}{c|}{VICReg}&77.73&44.58&74.06&26.20&81.38&53.46&77.45&34.17\\
\multicolumn{2}{c|}{TriMix(ours)}&\textbf{86.35}&\textbf{54.01}&\textbf{81.59}&34.66&\textbf{88.39}&\textbf{63.37}&\textbf{87.06}&45.15\\
\hline
\end{tabular}}
\end{table*}
\begin{table*}[!t] 
\caption{Top-1 accuracy($\%$) of the semi-supervised learning results (with 100 fine tuning epochs) using $1\%$, $10\%$ and
$100\%$ training examples on the benchmark datasets. The best results are in bold. Our method outperforms three out of 4 datasets at lower data regimes (at 1\% and 10\%), while it obtains the second-best results at 100\% data split.}
\label{tablesemi}
\centering
\resizebox{1\textwidth}{!}{
\begin{tabular}{l|l|c c c c| c c c c| c c c c}
\hline
\multicolumn{2}{c}{}& \multicolumn{4}{c|}{\textbf{1}\%}&\multicolumn{4}{c}{\textbf{10}\%}&\multicolumn{4}{c}{\textbf{100}\%}\\
\hline
\multicolumn{2}{c|}{Method}&CIFAR10&CIFAR100&STL10&TinyImageNet& CIFAR10&CIFAR100&STL10&TinyImageNet& CIFAR10&CIFAR100&STL10&TinyImageNet\\
\hline 
\hline
\multicolumn{2}{c|}{SimCLR}&79.03&29.94&65.23&16.14&86.16&51.65&79.98&35.5&\textbf{92.68}&\textbf{70.09}&88.94&\textbf{55.36}\\
\multicolumn{2}{c|}{BYOL}&77.27&28.04&61.05&15.68&85.01&50.36&79.03&34.13&92.29&69.56&88.74&54.26\\
\multicolumn{2}{c|}{Barlow Twins}&79.19&33.14&63.82&\textbf{20.28}&86.55&54.86&80.15&\textbf{39.83}&92.00&69.15&88.95&54.62\\
\multicolumn{2}{c|}{VICReg}&70.07&22.51&53.79&12.23&81.94&45.93&71.93&31.35&91.97&69.18&85.41&52.38\\
\multicolumn{2}{c|}{TriMix(ours)}&\textbf{81.03}&\textbf{34.08}&\textbf{66.80}&19.32&\textbf{87.56}&\textbf{56.23}&\textbf{81.08}&39.52&92.16&69.72&\textbf{89.54}&54.89\\
\hline
\end{tabular}}
\end{table*}
Our method obtains the best top-1 accuracy of $88.39\%$, $63.37\%$, and $87.06\%$ for the linear evaluation on CIFAR10, CIFAR100, and STL10 datasets, respectively, which are better than all baseline methods. Note that the KNN results reveal the same superiority of our approach over the baselines. On the other hand, TriMix achieves the second-best results and is on par with the Barlow Twins in the linear evaluation for the Tiny ImageNet dataset.
For more illustration, we draw the losses during the training of our method on TinyImage in Fig.\ref{figLoss}. The curves show that our losses (virtual embeddings and consistency) are beneficial and contribute significantly to the training. Despite that, we notice negligible oscillations in the consistency loss, which is attributed to the complexity of its task. However, its overall curve decreases during the training.
The above results demonstrate the importance of utilizing the manifold embeddings augmentations and the self-consistency tasks to achieve outstanding results in self-supervised learning. \\  

\noindent\textbf{\textit{Semi-supervised Evaluation (Natural images).}}
In this experiment, our method and the baselines were fine-tuned on subsets of $1\%$, $10\%$, and $100\%$ of the benchmark datasets for semi-supervised learning. We use a learning rate of $1\times10^{-3}$, weight decay of $1\times10^{-6}$, a momentum of $0.9$, and a batch size of $256$ for 100 epochs. The obtained semi-supervised results are reported in Table~\ref{tablesemi}. Our approach achieves the best results for the data splits at $1\%$ and $10\%$ on 3 out of 4 datasets. Yet, our method gets the second-best results for the $100\%$ of the data split. This experiment shows the effectiveness of our strategy in a lower data regime.\\

\noindent\textbf{\textit{KNN and Linear Evaluations (Medical images).}}
\begin{table*}[!t] 
\caption{Top-1 accuracy($\%$) of TriMix and the baselines for KNN (with 200-epoch pretraining) and linear evaluations ( with 100-epoch supervised training) on the medical datasets. Best results are in bold. Our method outperforms other methods in the medical datasets in the KNN and linear evaluations, confirming the results in the previous experiments.}
\label{tablelinear_med}
\centering
\resizebox{1\textwidth}{!}{
\begin{tabular}{l|l|c c c c| c c c c}
\hline
\multicolumn{2}{c}{}& \multicolumn{4}{c|}{\textbf{KNN}}&\multicolumn{4}{c}{\textbf{Linear}}\\
\hline
\multicolumn{2}{c|}{Method}&PathMNIST&DermaMNIST&OCTMNIST&BloodMNIST&PathMNIST&DermaMNIST&OCTMNIST&BloodMNIST\\
\hline 
\hline
\multicolumn{2}{c|}{SimCLR}&91.56&71.87&69.90&89.42&92.69&73.41&74.5&92.90\\
\multicolumn{2}{c|}{BYOL}&91.50&71.16&70.00&86.14&92.72&72.47&75.80&90.73\\
\multicolumn{2}{c|}{Barlow Twins}&91.36&71.68&71.20&87.89&92.42&73.36&\textbf{77.50}&92.28\\
\multicolumn{2}{c|}{VICReg}&90.40&70.77&69.50&86.14&90.87&72.24&72.39&89.16\\
\multicolumn{2}{c|}{TriMix(ours)}&\textbf{91.92}&\textbf{72.07}&\textbf{72.30}&\textbf{89.92}&\textbf{92.88}&\textbf{73.82}&76.60&\textbf{93.16}\\
\hline
\end{tabular}}
\end{table*}
Thus far, we have shown the performance of TriMix in four natural datasets. In the following experiments, we validate our method on four publicly available medical datasets from MedMNIST. Note that we kept the same setup from previous experiments. In addition, we evaluated the pre-trained representations using a supervised linear classifier on the frozen representations from the pre-training step on the medical data. The results are reported in Table \ref{tablelinear_med}. 
Starting with the KNN results, TriMix outperforms all other methods in the four medical data with a classification accuracy of $91.92$, $72.07$, $72.30$, and $89.92$ for PathMNIST, DermaMNIST, OCTMNIST, and BloodMNIST, respectively, with improvement between $0.2\%$ and $1.1\%$ better than the second-best models, confirming the results in the previous experiments. The same superiority also is found in the results of the linear evaluation. For example,  except for OCT images, our approach outperforms all methods with accuracy reaching $93.16$ in blood cell microscopic images. \\ 

\noindent\textbf{\textit{Semi-supervised Evaluation (Medical images).}}
\begin{table*}[!t] 
\caption{Top-1 accuracy($\%$) of the semi-supervised learning results (with 100 fine tuning epochs) using $1\%$, $10\%$ and
$100\%$ training examples on 4 medical datasets. Best results are in bold. In general, our method shows the best results in all data split.}
\label{tablesemi_med}
\centering
\resizebox{1\textwidth}{!}{
\begin{tabular}{l|l|c c c c| c c c c| c c c c}
\hline
\multicolumn{2}{c}{}& \multicolumn{4}{c|}{\textbf{1}\%}&\multicolumn{4}{c}{\textbf{10}\%}&\multicolumn{4}{c}{\textbf{100}\%}\\
\hline
\multicolumn{2}{c|}{Method}&PathMNIST&DermaMNIST&OCTMNIST&BloodMNIST&PathMNIST&DermaMNIST&OCTMNIST&BloodMNIST&PathMNIST&DermaMNIST&OCTMNIST&BloodMNIST\\
\hline 
\hline
\multicolumn{2}{c|}{SimCLR}&90.47&66.88&77.30&84.68&91.95&71.22&80.69&\textbf{92.15}&\textbf{93.50}&75.86&\textbf{83.89}&95.81\\
\multicolumn{2}{c|}{BYOL}&89.07&67.11&78.40&80.64&92.09&71.32&80.01&91.25&93.34&75.31&83.60&\textbf{96.11}\\
\multicolumn{2}{c|}{Barlow Twins}&89.90&67.13&78.30&82.61&92.08&70.97&80.20&91.34&92.54&75.86&82.30&95.90\\
\multicolumn{2}{c|}{VICReg}&89.77&66.98&79.30&78.54&91.61&\textbf{71.37}&80.59&88.97&93.38&76.00&81.50&95.32\\
\multicolumn{2}{c|}{TriMix(ours)}&\textbf{90.81}&\textbf{67.23}&\textbf{78.80}&\textbf{84.89}&\textbf{92.15}&\textbf{71.37}&\textbf{80.78}&\textbf{92.15}&\textbf{93.50}&\textbf{76.21}&81.90&95.73\\
\hline
\end{tabular}}
\end{table*}
The semi-supervised learning for the medical data is presented in Table \ref{tablesemi_med}. As in the previous setting,  we fine-tune the pre-trained models on subsets of $1\%$, $10\%$, and $100\%$ of datasets. The table clearly shows that our method achieves the best results in all datasets at all data splits. Exception from that is the results of OCT and Blood datasets at $100\%$ of the data. For instance, the accuracy of our approach at $1\%$, $10\%$, and $100\%$ of the data are $90.81$, $92.15$, and $93.50$ in PathMNIST, $67.23$, $71.37$, and $76.21$ in DermaMNIST, $78.80$, $80.78$, and $81.90$ in OCTMNIST, and $84.89$, $92.15$, and $95.73$ in BloodMNIST, respectively.  
These experiments in both medical and natural datasets reveal the generalizability and applicability of our method to various types of images in self/semi-supervised settings. Nearly in all these experiments, TriMix has superiority over others. Still, another critical issue is to investigate our method in a transfer learning setting.  \\      
\begin{table*}[!t] 
\caption{Top-1 accuracy($\%$) of transfer learning experiments of the pre-trained models on natural data to the medical data. The best results are in bold. Our learned representations capture more beneficial information and generalize better than the remaining approaches.}
\label{tabletrans}
\centering
\resizebox{1\textwidth}{!}{
\begin{tabular}{l|l|c c c c| c c c c}
\hline
\multicolumn{2}{c}{}& \multicolumn{4}{c|}{\textbf{CIFAR10}}&\multicolumn{4}{c}{\textbf{CIFAR100}}\\
\hline
\hline
\multicolumn{2}{c|}{Method}&PathMNIST&DermaMNIST&OCTMNIST&BloodMNIST&PathMNIST&DermaMNIST&OCTMNIST&BloodMNIST\\
\hline 
\multicolumn{2}{c|}{SimCLR}&77.17&71.57&63.90&77.81&76.43&71.57&60.50&76.23\\
\multicolumn{2}{c|}{BYOL}&78.19&72.21&65.00&79.42&80.78&71.52&57.30&75.85\\
\multicolumn{2}{c|}{Barlow Twins}&78.30&71.67&62.70&\textbf{80.70}&77.60&\textbf{73.01}&58.40&\textbf{80.35}\\
\multicolumn{2}{c|}{VICReg}&77.47&71.77&63.50&77.19&80.87&71.22&\textbf{65.10}&76.79\\
\multicolumn{2}{c|}{TriMix(ours)}&\textbf{78.43}&\textbf{72.27}&\textbf{67.20}&78.49&\textbf{80.92}&71.17&57.40&78.84\\
\hline
\hline
\multicolumn{2}{c}{}& \multicolumn{4}{c|}{\textbf{STL10}}&\multicolumn{4}{c}{\textbf{TinyImageNet}}\\
\hline
\hline
\multicolumn{2}{c|}{Method}&PathMNIST&DermaMNIST&OCTMNIST&BloodMNIST&PathMNIST&DermaMNIST&OCTMNIST&BloodMNIST\\
\hline 
\multicolumn{2}{c|}{SimCLR}&74.27&70.22&40.90&61.47&77.43&70.92&46.00&69.48\\
\multicolumn{2}{c|}{BYOL}&72.93&69.52&\textbf{57.90}&\textbf{67.72}&77.42&70.97&52.10&\textbf{73.98}\\
\multicolumn{2}{c|}{Barlow Twins}&\textbf{77.40}&70.12&50.20&67.17&76.50&71.06&\textbf{56.20}&73.77\\
\multicolumn{2}{c|}{VICReg}&71.10&70.02&44.60&60.07&72.86&71.02&47.50&64.42\\
\multicolumn{2}{c|}{TriMix(ours)}&76.56&\textbf{70.32}&49.30&63.84&\textbf{77.77}&\textbf{71.87}&51.20&72.46\\
\hline
\end{tabular}}
\end{table*}

\noindent\textbf{\textit{Transfer learning from natural to medical data.}}
This experiment aims to build linear classifiers on top of fixed representations of the pre-trained models on the natural images. Then, we fine-tune these models on the four medical datasets simulating the transfer learning setting as in the literature, which resulted in 80 models shown in Table \ref{tabletrans}. One can notice that our approach generalizes better when using a pre-trained model on CIFAR10 and TinyImage. Yet, Barlow Twins works better using CIFAR100, and BYOL outperforms others using STL10. Generally, TriMix is among the best models regardless of the used dataset. This experiment reveals that our learned representations capture more beneficial information and generalize better than the remaining approaches.            
\subsection{Comprehensive analysis of TriMix}
\label{exp_ana}
\begin{figure*}[!t]
\centering
\includegraphics[width=\textwidth]{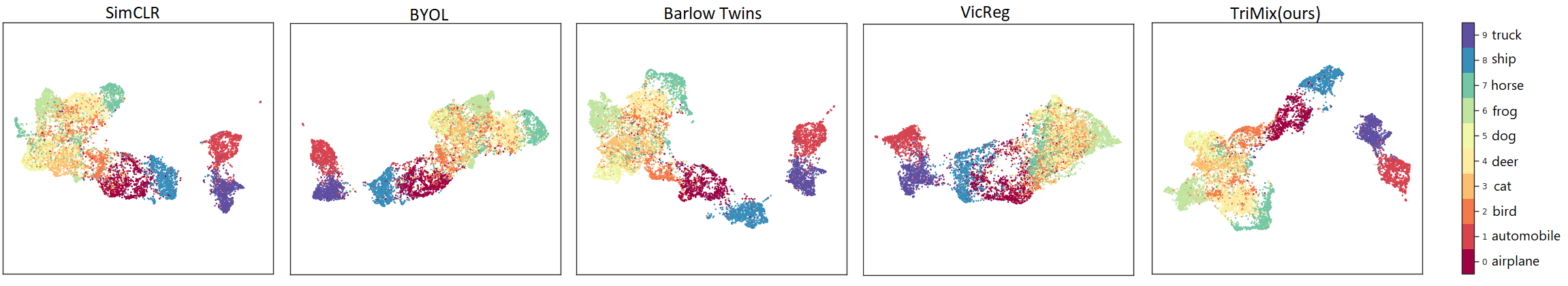}
\caption{2D UMAP projection of CIFAR10 testing dataset using different self-supervised learning methods. Our method (TriMix) is better and less noisy in clustering the ten classes than the remaining methods.}
\label{figUMAP}
\end{figure*}

\noindent\textbf{\textit{Clustering the learned representations $Y$ analysis.}}
To gain more insights into the effect of our approach on the learned representations and realize the differences between ours and the baselines, we visualize the learned representations using UMAP~\cite{mcinnes2018umap}, an open-source library for dimensionality reduction. As an illustration, we cluster the learned representations for the CIFAR10 testing dataset in Figure~\ref{figUMAP}.   
It is noticed that our method clusters the ten classes more precisely than the others. For example, the classes of "truck" and "automobile" are more compact by ours than the others, while the classes of "ship" and "airplane" are less overlapped with each other. On the other hand, the animals' classes (\ie bird, cat, deer, dog, frog, and horse) represent a challenge to all models; nevertheless, our method clusters them with lower noises. \\

\begin{table}[!t] 
\caption{Objective function components analysis. Linear evaluation on the CIFAR10 and STL10 data. Our contributions enhance the baseline while combining them boosts the performance significantly.}
\label{tableabla}
\centering
\resizebox{0.5\textwidth}{!}{
\begin{tabular}{l|c|c}
\hline
Configuration&CIFAR10\\
\hline
\hline
Barlow Twins&86.29\\
\hline
\textbf{TriMix}: Virtual embeddings loss only&87.16\\
\textbf{TriMix}: Self-consistency loss only&87.32\\
\textbf{TriMix}: Virtual embeddings loss + self-consistency loss&87.74\\
\textbf{TriMix}: Virtual embeddings loss + self-consistency loss + feature norms&\textbf{88.39}\\
\hline 
\end{tabular}
}
\end{table}

\noindent\textbf{\textit{Objective function components analysis.}}
This experiment investigates the effect of our main contributions. We train different versions of TriMix with(out) virtual embeddings and self-consistency terms on the CIFAR10 dataset for 200 epochs. Then, we conduct a linear evaluation for 100 epochs and register the results in Table~\ref{tableabla}. First, it is shown that utilizing either of our contributions enhances total accuracy with $87.16$ and $87.32$ when using the virtual and self-consistency losses, respectively. While combining both terms adds more enhancements up to $87.74$. Finally, joining all contributions, including the features normalization, plays a significant role in the overall performance at $88.39$. \\

\noindent\textbf{\textit{Batch and projector sizes analysis.}}
\begin{figure*}[!t]
\centering
\includegraphics[width=1\textwidth]{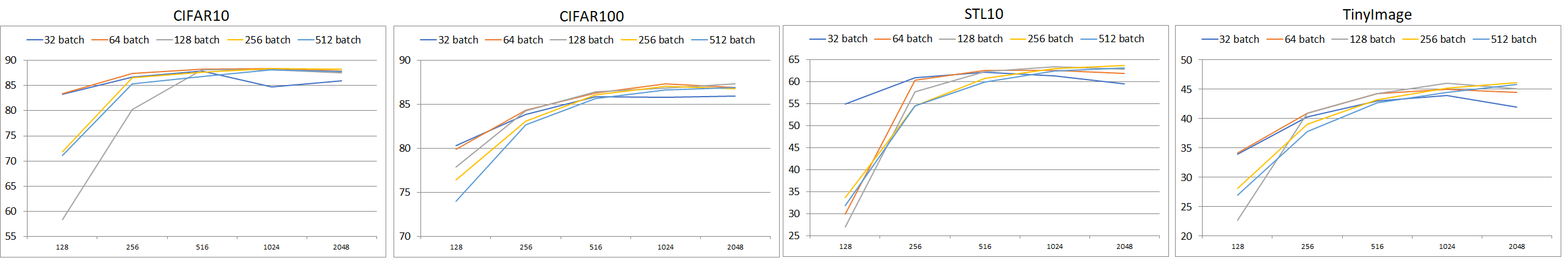}
\caption{Top-1 accuracy($\%$) of TriMix when changing the batch size and features dimension. (i) Batch size: increasing the size of the features enhances the results. (ii) Features dimension: smaller batches work better with smaller dimensions and vice versa.}
\label{figBatchFeat}
\end{figure*}  
Further, we test the effect of changing the batch size and projector dimension on TriMix. Specifically, we investigate batches of size \{32, 64, 128, 256, 512\}, with features dimensions of \{128, 256, 512, 1024, 2048\} on CIFAR10/100, STL10, and TinyImage. The results are curves for 100 models, presented in Fig.\ref{figBatchFeat}. In general, the accuracy curves show that for fixed batch size, using more extensive features achieves better results, which agrees with the literature. Also, for most models, no significant difference in the accuracy for dimensions 1024 and 2048. Further, for fixed feature size, the smaller batches work better with smaller dimensions, while bigger batches favor bigger dimensions. \\

\noindent\textbf{\textit{Virtual data analysis.}}
\begin{figure*}[!t]
\centering
\includegraphics[width=\textwidth]{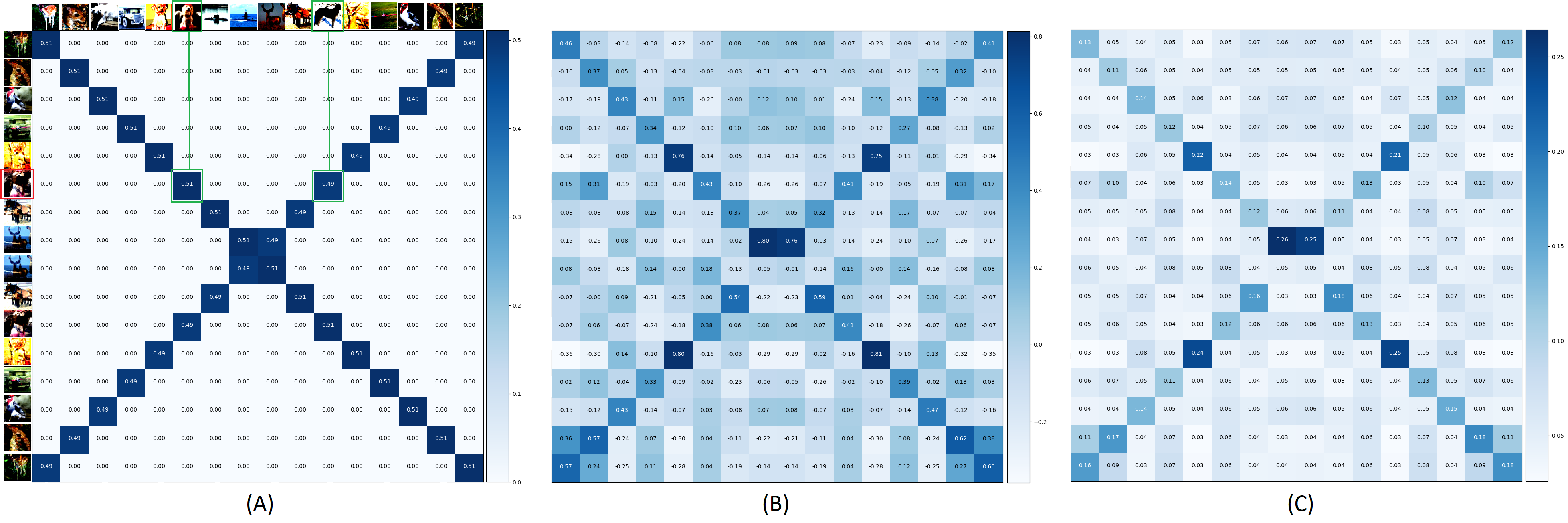}
\caption{Mixed data analysis. (A) sample ground truth matrix. We depict the corresponding batch in the above raw. To the left of the matrix, we display the virtual images. For example, the image in the red rectangle is generated from the images in the green rectangles. (B) and (C) show the predicted matrix before and after softmax operation, respectively. By raining on this task, our method gains more information from the images, and hence, better performance.}
\label{figMix}
\end{figure*} 
This experiment aims to realize how our method decomposes and regresses the augmented data from its mixture. In Fig.\ref{figMix}, we present sample images from the STL10 dataset. Fig.\ref{figMix}.(A) shows a sample ground truth matrix ($\mathcal{GT}$), which is created at each epoch, see Eq.(\ref{eq_L_vrt}). To illustrate, we also show the corresponding batch in the above raw. To the left of the matrix, we display the augmented images. For example, the one in the red rectangle is generated by mixing the two images from the original batch, \ie, the images in green rectangles in the above raw, using $\lambda$ and $1-\lambda$ (0.51 and 0.49, respectively in this sample). Figures \ref{figMix}.(B) and \ref{figMix}.(C), on the other hand, show the predicted matrix before and after softmax operation, which are corresponding to Eq.(\ref{eq_CorMatM}) and Eq.(\ref{eq_soft}), respectively. Our method attempts to anticipate the ground truth matrix and decomposes the mixed images into the original parts, as shown in Fig.\ref{figMix} (B). Thus, by training on this auxiliary task, our method learns to distinguish between different images and gains more information, which boosts performance. Notice that the predicted matrix, in Fig.\ref{figMix} (B), contains some noises, \ie, negative values. Yet, after applying the softmax operation, Eq.(\ref{eq_soft}), the model produces more stable predictions and hence better results.\\

\noindent\textbf{\textit{TriMix variants analysis.}}
Our initial setting depends on $Z$ to train the model. While in this experiment, we explore alternative ways of implementing our loss functions. Specifically, we attempt to use the hidden feature representation $Y$ in our objective functions. The first alternative is to define virtual embeddings loss and self-consistency loss on feature representation $Y$. The second choice is to optimize virtual embeddings loss on embeddings $Z$ and self-consistency loss on hidden feature $Y$. Eventually, we compared both experiments with our baseline model and reported the results in Table \ref{tablevar}.
In short, our initial choice for the overall loss function that depends on $Y$ is still the most reasonable approach. However, using our losses on different data types, \ie, virtual embeddings loss on embeddings $Z$ and self-consistency loss on hidden feature $Y$ achieve the lowest accuracy. That is related to the fact that each term works on incompatible features (\ie, $Y$ and $Z$), which is confusing our objective function.
\begin{table}[!t] 
\caption{TriMix: Loss function based on feature representation $Y$ analysis. Linear evaluation on the CIFAR10. Our initial setting is the most valid option. While using different features in both terms confuses the model.}
\label{tablevar}
\centering
\resizebox{0.5\textwidth}{!}{
\begin{tabular}{l|c|c}
\hline
Configuration&Accuracy\\
\hline
\hline
Virtual embeddings loss on $Y$ + self-consistency loss on $Y$ &  86.32 \\
Virtual embeddings loss on $Z$ + self-consistency loss on $Y$ &  85.37 \\
Virtual embeddings loss on $Z$ + self-consistency loss on $Z$ (Baseline) &  \textbf{88.39} \\
\hline 
\end{tabular}
}
\end{table}

\section{Discussion}
\label{disc}
Our method; TriMix, proposed i) virtual embeddings loss: first, we augment the network with novel embeddings generated from the original ones, then we train the model to decompose these virtual embeddings to their original components, 2) self-consistency term that enforces the consistency between the virtual and the original data. When compared with the other approaches, TriMix showed the best results in the vast majority of our experiments.\\

\noindent\textbf{\textit{Applicability and Transferability.}}
Our method has shown to be effective in all experiments beating recent SSL methods in the majority of the tasks. First, we have demonstrated the applicability of our approach to eight public datasets, including natural and medical images. Also, our strategy was found very beneficial in semi-supervised learning settings, especially at lower data regimes \ie $1\%$ and $10\%$, where the models suffer from scarcity in the labeled data. That corresponds to the need, in this particular setting,  for additional and novel representations to augment the model with new training data. Even though no specific approach was dominated in the transfer learning experiments, due to the complexity of transferring the pre-trained models, using any natural dataset, to all medical data and achieving SOTA performance. Still, among all pre-trained models,  ours were the most successful. 
The above results shed light on manifold augmentation and self-consistency in self-supervised learning.\\  

\noindent\textbf{\textit{Manifold and Hidden Embeddings Augmentation.}} 
The current self-supervised learning methods heavily depend on two views or augmentations of the input data to train the network. While this is an essential step for any successful self-supervised approach, none of the previous efforts have investigated the augmentation at the manifold or hidden representations. We have shown in this work that attaching the manifold augmentation to the training boosts performance. Our augmentation methodology depends on mixing the original embeddings with random percentages while training the model to predict these percentages and decompose the augmented data to the original elements. 
Analyzing the hidden representations shows that our method is better at clustering the classes with fewer noises, which justifies its top performance over other baselines. Still, an important question we have addressed in the paper was where to inject this function. We have shown that placing the two losses at the hidden embeddings, \ie $Z$, achieved the best results. That is attributed to the homogeneity of used data in both terms. On the other hand, placing the two losses at different locations, \ie embedding $Z$ or hidden representations $Y$, might confuse the model and reduce its performance. 
While we kept our augmentation methodology simple, one can investigate more sophisticated augmentation approaches or inject more than extra tasks. Anyway, model failure is one of the most challenging tasks in that situation.\\ 

\noindent\textbf{\textit{Interpretability.}}
TriMix performance is related to the training strategy we introduced. Our auxiliary task serves two purposes. First, augment the model with novel data generated from the mixup operation at the manifold layers. It is known that data augmentation plays a fundamental role in the performance of any deep learning method. Second, train the model to distinguish between the images by predicting the mixing ratio used to generate the new data. For illustration, consider a case when mixing a malignant sample with a benign one. Now, when we train our model to decompose the combined image to the original ones. The model implicitly learns the characteristic of each class. Thus, more information is being realized by achieving this task.

\section{Conclusion}
\label{conc}
In this paper, we have proposed TriMix, a self-supervised learning method that introduces virtual embeddings and self-consistency in training. While the current works depend on data augmentations at the input space in Siamese-based architectures, our approach proposed an auxiliary task that generates new virtual data through linear interpolation of hidden embeddings. Besides providing new data, our training strategy is to learn the model to predict the mixing factor used in generating the data such that the model can distinguish between the images and decompose the mixed data to the original components. Further, we propose a loss term to force self-consistency in the data. We have shown the applicability of our method on eight public datasets consisting of natural and medical images with improvements reaching $2.71\%$ and $0.41\%$ better than the second-best models for both data types, respectively. Moreover, TriMix demonstrated superior performance in the low amount of data, \ie $1\%$ and $10\%$, of the semi-supervised learning setting while on par with best models when using $100\%$ of the data. Although none of the methods excels in all transfer learning experiments, our pre-trained models showed the best accuracy. Our strategy highlights the importance of the embeddings augmentations and the additional tasks to achieve leading results. We opt for a simple augmentation methodology to avoid any potential model collapsing, yet, studying different ones, including sophisticated methods, could be a future research direction. Additionally, we build a ground truth matrix based on random $\lambda$ at the beginning of each epoch. However, we could investigate a more intelligent and adaptive way that evolved during the training.  

\section*{Acknowledgments}
T.B. is financially supported by the German Academic Exchange Service (DAAD)

\printbibliography

@article{zbontar2021barlow,
  title={Barlow twins: Self-supervised learning via redundancy reduction},
  author={Zbontar, Jure and Jing, Li and Misra, Ishan and LeCun, Yann and Deny, St{\'e}phane},
  journal={arXiv preprint arXiv:2103.03230},
  year={2021}
}

@article{tian2021divide,
  title={Divide and Contrast: Self-supervised Learning from Uncurated Data},
  author={Tian, Yonglong and Henaff, Olivier J and Oord, Aaron van den},
  journal={arXiv preprint arXiv:2105.08054},
  year={2021}
}

@article{chen2020big,
  title={Big self-supervised models are strong semi-supervised learners},
  author={Chen, Ting and Kornblith, Simon and Swersky, Kevin and Norouzi, Mohammad and Hinton, Geoffrey},
  journal={arXiv preprint arXiv:2006.10029},
  year={2020}
}

@article{grill2020bootstrap,
  title={Bootstrap your own latent: A new approach to self-supervised learning},
  author={Grill, Jean-Bastien and Strub, Florian and Altch{\'e}, Florent and Tallec, Corentin and Richemond, Pierre H and Buchatskaya, Elena and Doersch, Carl and Pires, Bernardo Avila and Guo, Zhaohan Daniel and Azar, Mohammad Gheshlaghi and others},
  journal={arXiv preprint arXiv:2006.07733},
  year={2020}
}

@article{caron2020unsupervised,
  title={Unsupervised learning of visual features by contrasting cluster assignments},
  author={Caron, Mathilde and Misra, Ishan and Mairal, Julien and Goyal, Priya and Bojanowski, Piotr and Joulin, Armand},
  journal={arXiv preprint arXiv:2006.09882},
  year={2020}
}

@article{devlin2018bert,
  title={Bert: Pre-training of deep bidirectional transformers for language understanding},
  author={Devlin, Jacob and Chang, Ming-Wei and Lee, Kenton and Toutanova, Kristina},
  journal={arXiv preprint arXiv:1810.04805},
  year={2018}
}

@article{yang2019xlnet,
  title={Xlnet: Generalized autoregressive pretraining for language understanding},
  author={Yang, Zhilin and Dai, Zihang and Yang, Yiming and Carbonell, Jaime and Salakhutdinov, Russ R and Le, Quoc V},
  journal={Advances in neural information processing systems},
  volume={32},
  year={2019}
}

@article{xiong2016achieving,
  title={Achieving human parity in conversational speech recognition},
  author={Xiong, Wayne and Droppo, Jasha and Huang, Xuedong and Seide, Frank and Seltzer, Mike and Stolcke, Andreas and Yu, Dong and Zweig, Geoffrey},
  journal={arXiv preprint arXiv:1610.05256},
  year={2016}
}

@article{oord2018representation,
  title={Representation learning with contrastive predictive coding},
  author={Oord, Aaron van den and Li, Yazhe and Vinyals, Oriol},
  journal={arXiv preprint arXiv:1807.03748},
  year={2018}
}

@article{bromley1993signature,
  title={Signature verification using a" siamese" time delay neural network},
  author={Bromley, Jane and Guyon, Isabelle and LeCun, Yann and S{\"a}ckinger, Eduard and Shah, Roopak},
  journal={Advances in neural information processing systems},
  volume={6},
  year={1993}
}

@inproceedings{he2020momentum,
  title={Momentum contrast for unsupervised visual representation learning},
  author={He, Kaiming and Fan, Haoqi and Wu, Yuxin and Xie, Saining and Girshick, Ross},
  booktitle={Proceedings of the IEEE/CVF Conference on Computer Vision and Pattern Recognition},
  pages={9729--9738},
  year={2020}
}

@inproceedings{chen2020simple,
  title={A simple framework for contrastive learning of visual representations},
  author={Chen, Ting and Kornblith, Simon and Norouzi, Mohammad and Hinton, Geoffrey},
  booktitle={International conference on machine learning},
  pages={1597--1607},
  year={2020},
  organization={PMLR}
}

@inproceedings{verma2019manifold,
  title={Manifold mixup: Better representations by interpolating hidden states},
  author={Verma, Vikas and Lamb, Alex and Beckham, Christopher and Najafi, Amir and Mitliagkas, Ioannis and Lopez-Paz, David and Bengio, Yoshua},
  booktitle={International Conference on Machine Learning},
  pages={6438--6447},
  year={2019},
  organization={PMLR}
}

@article{bdair2020roam,
  title={ROAM: Random Layer Mixup for Semi-Supervised Learning in Medical Imaging},
  author={Bdair, Tariq and Wiestler, Benedikt and Navab, Nassir and Albarqouni, Shadi},
  journal={arXiv preprint arXiv:2003.09439},
  year={2020}
}

@article{zhang2017mixup,
  title={mixup: Beyond empirical risk minimization},
  author={Zhang, Hongyi and Cisse, Moustapha and Dauphin, Yann N and Lopez-Paz, David},
  journal={arXiv preprint arXiv:1710.09412},
  year={2017}
}

@article{hua2021feature,
  title={On Feature Decorrelation in Self-Supervised Learning},
  author={Hua, Tianyu and Wang, Wenxiao and Xue, Zihui and Wang, Yue and Ren, Sucheng and Zhao, Hang},
  journal={arXiv preprint arXiv:2105.00470},
  year={2021}
}

@inproceedings{huang2018decorrelated,
  title={Decorrelated batch normalization},
  author={Huang, Lei and Yang, Dawei and Lang, Bo and Deng, Jia},
  booktitle={Proceedings of the IEEE Conference on Computer Vision and Pattern Recognition},
  pages={791--800},
  year={2018}
}

@inproceedings{deng2009imagenet,
  title={Imagenet: A large-scale hierarchical image database},
  author={Deng, Jia and Dong, Wei and Socher, Richard and Li, Li-Jia and Li, Kai and Fei-Fei, Li},
  booktitle={2009 IEEE conference on computer vision and pattern recognition},
  pages={248--255},
  year={2009},
  organization={Ieee}
}

@article{hjelm2018learning,
  title={Learning deep representations by mutual information estimation and maximization},
  author={Hjelm, R Devon and Fedorov, Alex and Lavoie-Marchildon, Samuel and Grewal, Karan and Bachman, Phil and Trischler, Adam and Bengio, Yoshua},
  journal={arXiv preprint arXiv:1808.06670},
  year={2018}
}

@inproceedings{wu2018unsupervised,
  title={Unsupervised feature learning via non-parametric instance discrimination},
  author={Wu, Zhirong and Xiong, Yuanjun and Yu, Stella X and Lin, Dahua},
  booktitle={Proceedings of the IEEE conference on computer vision and pattern recognition},
  pages={3733--3742},
  year={2018}
}

@inproceedings{henaff2020data,
  title={Data-efficient image recognition with contrastive predictive coding},
  author={Henaff, Olivier},
  booktitle={International Conference on Machine Learning},
  pages={4182--4192},
  year={2020},
  organization={PMLR}
}

@inproceedings{tian2020contrastive,
  title={Contrastive multiview coding},
  author={Tian, Yonglong and Krishnan, Dilip and Isola, Phillip},
  booktitle={Computer Vision--ECCV 2020: 16th European Conference, Glasgow, UK, August 23--28, 2020, Proceedings, Part XI 16},
  pages={776--794},
  year={2020},
  organization={Springer}
}

@article{bachman2019learning,
  title={Learning representations by maximizing mutual information across views},
  author={Bachman, Philip and Hjelm, R Devon and Buchwalter, William},
  journal={arXiv preprint arXiv:1906.00910},
  year={2019}
}

@article{srinivas2020curl,
  title={Curl: Contrastive unsupervised representations for reinforcement learning},
  author={Srinivas, Aravind and Laskin, Michael and Abbeel, Pieter},
  journal={arXiv preprint arXiv:2004.04136},
  year={2020}
}

@article{bardes2021vicreg,
  title={Vicreg: Variance-invariance-covariance regularization for self-supervised learning},
  author={Bardes, Adrien and Ponce, Jean and LeCun, Yann},
  journal={arXiv preprint arXiv:2105.04906},
  year={2021}
}

@article{asano2019self,
  title={Self-labelling via simultaneous clustering and representation learning},
  author={Asano, Yuki Markus and Rupprecht, Christian and Vedaldi, Andrea},
  journal={arXiv preprint arXiv:1911.05371},
  year={2019}
}

@inproceedings{caron2018deep,
  title={Deep clustering for unsupervised learning of visual features},
  author={Caron, Mathilde and Bojanowski, Piotr and Joulin, Armand and Douze, Matthijs},
  booktitle={Proceedings of the European Conference on Computer Vision (ECCV)},
  pages={132--149},
  year={2018}
}

@inproceedings{xie2016unsupervised,
  title={Unsupervised deep embedding for clustering analysis},
  author={Xie, Junyuan and Girshick, Ross and Farhadi, Ali},
  booktitle={International conference on machine learning},
  pages={478--487},
  year={2016},
  organization={PMLR}
}

@inproceedings{chen2021exploring,
  title={Exploring simple siamese representation learning},
  author={Chen, Xinlei and He, Kaiming},
  booktitle={Proceedings of the IEEE/CVF Conference on Computer Vision and Pattern Recognition},
  pages={15750--15758},
  year={2021}
}

@article{cao2020parametric,
  title={Parametric instance classification for unsupervised visual feature learning},
  author={Cao, Yue and Xie, Zhenda and Liu, Bin and Lin, Yutong and Zhang, Zheng and Hu, Han},
  journal={arXiv preprint arXiv:2006.14618},
  year={2020}
}

@inproceedings{chaitanya2019semi,
  title={Semi-supervised and task-driven data augmentation},
  author={Chaitanya, Krishna and Karani, Neerav and Baumgartner, Christian F and Becker, Anton and Donati, Olivio and Konukoglu, Ender},
  booktitle={International conference on information processing in medical imaging},
  pages={29--41},
  year={2019},
  organization={Springer}
}

@article{eaton2018improving,
  title={Improving data augmentation for medical image segmentation},
  author={Eaton-Rosen, Zach and Bragman, Felix and Ourselin, Sebastien and Cardoso, M Jorge},
  year={2018}
}

@inproceedings{panfilov2019improving,
  title={Improving robustness of deep learning based knee mri segmentation: Mixup and adversarial domain adaptation},
  author={Panfilov, Egor and Tiulpin, Aleksei and Klein, Stefan and Nieminen, Miika T and Saarakkala, Simo},
  booktitle={Proceedings of the IEEE/CVF International Conference on Computer Vision Workshops},
  pages={0--0},
  year={2019}
}

@inproceedings{jung2019prostate,
  title={Prostate cancer segmentation using manifold mixup U-Net},
  author={Jung, Wonmo and Park, Sejin and Jung, Kyu-Hwan and Hwang, Sung Il},
  booktitle={International Conference on Medical Imaging with Deep Learning--Extended Abstract Track},
  year={2019}
}

@article{berthelot2019mixmatch,
  title={Mixmatch: A holistic approach to semi-supervised learning},
  author={Berthelot, David and Carlini, Nicholas and Goodfellow, Ian and Papernot, Nicolas and Oliver, Avital and Raffel, Colin},
  journal={arXiv preprint arXiv:1905.02249},
  year={2019}
}

@article{krizhevsky2009learning,
  title={Learning multiple layers of features from tiny images},
  author={Krizhevsky, Alex and Hinton, Geoffrey and others},
  year={2009},
  publisher={Citeseer}
}

@inproceedings{coates2011analysis,
  title={An analysis of single-layer networks in unsupervised feature learning},
  author={Coates, Adam and Ng, Andrew and Lee, Honglak},
  booktitle={Proceedings of the fourteenth international conference on artificial intelligence and statistics},
  pages={215--223},
  year={2011},
  organization={JMLR Workshop and Conference Proceedings}
}

@article{le2015tiny,
  title={Tiny imagenet visual recognition challenge},
  author={Le, Ya and Yang, Xuan},
  journal={CS 231N},
  volume={7},
  number={7},
  pages={3},
  year={2015}
}

@article{kingma2014adam,
  title={Adam: A method for stochastic optimization},
  author={Kingma, Diederik P and Ba, Jimmy},
  journal={arXiv preprint arXiv:1412.6980},
  year={2014}
}

@inproceedings{he2016deep,
  title={Deep residual learning for image recognition},
  author={He, Kaiming and Zhang, Xiangyu and Ren, Shaoqing and Sun, Jian},
  booktitle={Proceedings of the IEEE conference on computer vision and pattern recognition},
  pages={770--778},
  year={2016}
}

@article{mcinnes2018umap,
  title={Umap: Uniform manifold approximation and projection for dimension reduction},
  author={McInnes, Leland and Healy, John and Melville, James},
  journal={arXiv preprint arXiv:1802.03426},
  year={2018}
}

@article{medmnistv2,
    title={MedMNIST v2: A Large-Scale Lightweight Benchmark for 2D and 3D Biomedical Image Classification},
    author={Yang, Jiancheng and Shi, Rui and Wei, Donglai and Liu, Zequan and Zhao, Lin and Ke, Bilian and Pfister, Hanspeter and Ni, Bingbing},
    journal={arXiv preprint arXiv:2110.14795},
    year={2021}
}

@inproceedings{medmnistv1,
    title={MedMNIST Classification Decathlon: A Lightweight AutoML Benchmark for Medical Image Analysis},
    author={Yang, Jiancheng and Shi, Rui and Ni, Bingbing},
    booktitle={IEEE 18th International Symposium on Biomedical Imaging (ISBI)},
    pages={191--195},
    year={2021}
}
\end{document}